\def\year{2021}\relax
\documentclass[letterpaper]{article} 
\usepackage{aaai21}  
\usepackage{times}  
\usepackage{helvet} 
\usepackage{courier}  
\usepackage[hyphens]{url}  
\usepackage{graphicx} 
\usepackage{natbib}  
\usepackage{caption} 
\usepackage{booktabs}
\usepackage{amsmath}
\usepackage{amssymb}
\usepackage{algorithm}
\usepackage{algorithmic}
\usepackage{array}
\usepackage{subfig}

\title{STDI-Net: Spatial-Temporal Network with Dynamic Interval Mapping for Bike Sharing Demand Prediction}

\author{Weiguo Pian\textsuperscript{\rm 1}, Yingbo Wu\thanks{Corresponding author}\textsuperscript{\rm 1}, Ziyi Kou\textsuperscript{\rm 2}\\ 
\textsuperscript{\rm 1}Chongqing University\\ 
\textsuperscript{\rm 2}University of Notre Dame\\
\{pwg,wyb\}@cqu.edu.cn, zkou@nd.edu
}
\begin{document}

\maketitle

\begin{abstract}
As an economical and healthy mode of shared transportation, Bike Sharing System (BSS) develops quickly in many big cities. An accurate prediction method can help BSS schedule resources in advance to meet the demands of users, and definitely improve operating efficiencies of it. However, most of the existing methods for similar tasks just utilize spatial or temporal information independently. Though there are some methods consider both, they only focus on demand prediction in a single location or between location pairs. In this paper, we propose a novel deep learning method called Spatial-Temporal Dynamic Interval Network (STDI-Net). The method predicts the number of renting and returning orders of multiple connected stations in the near future by modeling joint spatial-temporal information. Furthermore, we embed an additional module that generates dynamical learnable mappings for different time intervals, to include the factor that different time intervals have a strong influence on demand prediction in BSS. Extensive experiments are conducted on the NYC Bike dataset, the results demonstrate the superiority of our method over existing methods.
\end{abstract}

\section{Introduction}
With the rapid development of sharing economy around the world, Bike Sharing System (BSS) has become more and more popular in recent years~\cite{bikesharing1,bikesharing2}. It provides people with a convenient and environment-friendly way of traveling. Users can rent a bike from a BSS station by some apps on their mobile phones and then return the bike to a station after completing their travels.

However, efficiently maintaining these systems is still challenging since the schedule and allocation of these transportation resources vary a lot depending on specific user requirements. For example, the number of rental orders on the morning of a day has an extremely imbalanced distribution between residential areas and commercial places. Therefore, a demand prediction method for adjustments of bikes in advance can improve the efficiency of BSS greatly.

\begin{figure}[t]
    \centering
    \includegraphics[width=.45\textwidth]{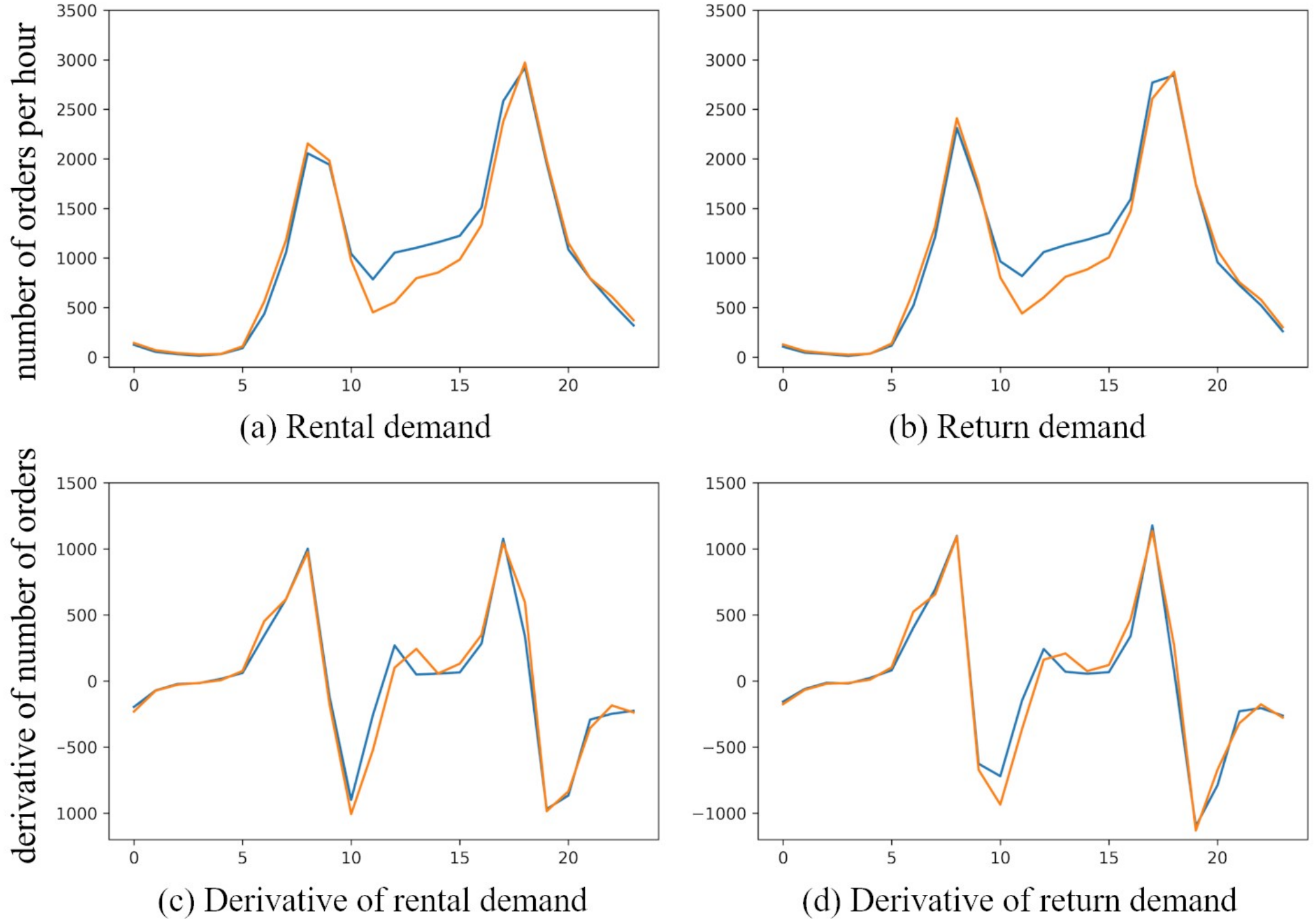}
    \caption{Number of orders and the rates of their changes during one day for both \textit{rent} and \textit{return} mode. The two lines with blue and orange color represent two single days in April 2014.}
    \label{fig:curve}
\end{figure}

To tackle this problem, there have been several methods proposed in recent years focusing on different prediction tasks. Besides some methods applying hand-crafted features~\cite{GaussianMixtureModel,DemandStreamingData,AdaptiveSeasonal}, one of the first deep learning methods was introduced by Wang et al.~\cite{DeepSD} who concatenated several related factors as inputs to predict the gap between taxi supply and demand via a non-linear MLP network. After that, Zhang et al.~\cite{AAAI2017} proposed a deep convolutional network named ST-ResNet to predict in-out traffic flow among different areas. However, both of them did not consider the temporal information hidden in the sequential data which is an important factor in transportation issues. Based on that, Yao et al.~\cite{AAAI2018} constructed a spatial-temporal model to predict various taxi demands. Moreover, they further created a graph embedding module to pass information among different areas. But their networks only consider a single area with its neighbors as inputs, thus obtains predicted results for different locations separately, which resulted in a serious lack of correlated spatial information on the global level.

Therefore, in our method, we construct a joint spatial-temporal network on a large scale area that contains hundreds of connected BSS stations in a long day hours. The network takes the number of both rental and returning orders of all stations in the past few hours as integrated inputs and predicts all of them in the near future together for once. By this way, the spatial correlation shared by all stations can be captured at the same level and same time, with global transportation information passing through each of them. Besides, the joint consideration of both operations for bikes, renting and returning, helps to maintain the sequential relation at each time interval. For the convolutional part, instead of applying the same filters for all features in different temporal indexes, we assign features in each index with one independent convolutional group. That is, we consider that indexes serve different roles in sequential data, which is far from enough to be captured by the same convolutional kernels. Compared with previous methods, our network can achieve much better performance with measurements of both accuracy and efficiency in demand prediction tasks.

Although all the previous methods have explored temporal information in a wide range, they all ignore an important factor that different time periods influence a lot on the change of demands. Based on that, we analyze the number of orders in BSS for each day and found in some periods, the orders increase or drop dramatically while for other times, no apparent fluctuation can be observed. As shown in Figure~\ref{fig:curve}, two colored lines are representing the change of orders in two single days and also their corresponding derivatives that further demonstrate the variety of demand changes in a continuous way. Therefore, we propose the dynamic interval module that takes different time intervals as inputs to improve the predictions of the main spatial-temporal network. Instead of applying a regular feature fusion for the outputs of the module, we get inspired by some few-shot learning methods~\cite{learnet,TAFENet} and directly assign the generated features as learnable parameters for the top layer that is responsible for final predictions in the main network. In such a learning framework, time intervals participate in the formulation of learning weights in a more straightforward way, which helps the whole model to learn a mapping that is adapted based on different time periods from the extracted spatial-temporal feature to the predicted demands.

\begin{figure}[t]
    \centering
    \includegraphics[width=.4\textwidth]{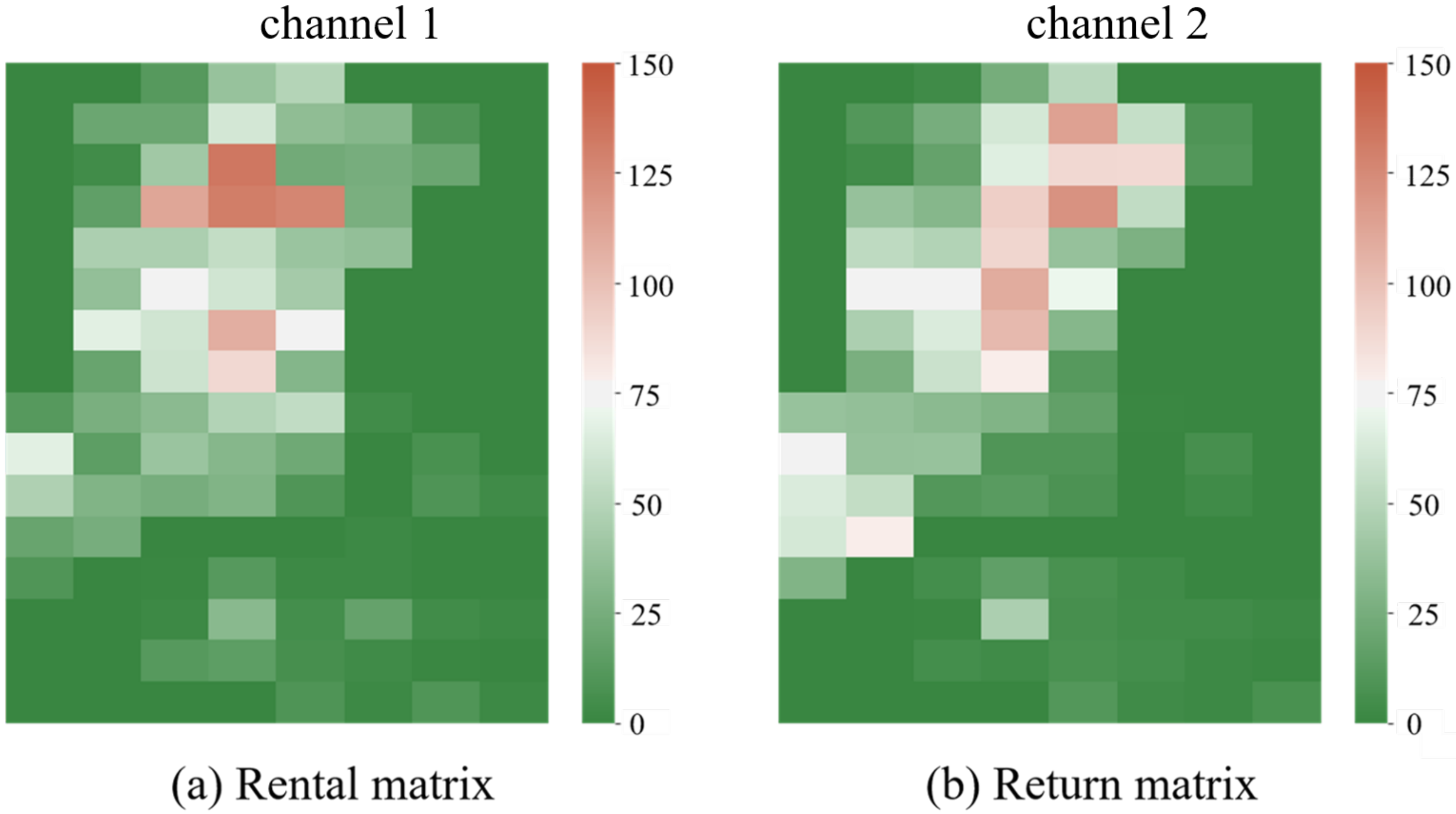}
    \caption{\textit{rental} and \textit{return} matrices as inputs for our joint spatial-temporal network. The sidebars for both matrices denote the relationship between colors and the number of orders.}
    \label{fig:demand_matrix}
\end{figure}

In summary, we collect our contribution into the following three folds:
\begin{itemize}
    \item We propose a joint spatial-temporal network with time-specific convolution layers to predict both renting and returning demand for all the stations in the BSS.
    \item We further propose a Dynamic Interval module that builds the relationship between different time intervals in a day to the learning representation that is assigned as learnable weights in the top regression layer.
    \item We conducted large scale experiments on the NYC Bike dataset. The result shows that our approach outperforms all other previous methods and several competitive baselines.
\end{itemize}

\section{Related Work}
Traffic prediction problems include many tasks, such as traffic flow prediction, destination prediction, demand prediction (our task), etc. The methods applied to these tasks are kind of similar. Essentially, they predict the data on future timestamps based on the historical one~\cite{DeepSD,AAAI2018,AAAI2017,DeepST}. Some traditional methods only rely on information in time series and regress final predictions. For instance, one of the most representative methods is Autoregressive Integrated Moving Average (ARIMA) which is widely used in traffic prediction problems~\cite{DemandStreamingData,AdaptiveSeasonal}. It takes continuous temporal information as inputs and regresses desired results. Besides, some other works included external context data, such as weather conditions and event information, to further improve the model's performance~\cite{realworld,UbiquitousUrbanData}.

Deep learning has been successfully used in a large number of problems, such as computer vision~\cite{ResidualNet,ImageNet}, which also widely used in traffic prediction.
Zhang et al.~\cite{DeepST} proposed a DNN-based model for predicting crowd flow. After that, they further introduced the residual connection originated from CNN-based networks~\cite{ResidualNet} for the same task~\cite{AAAI2017}. To utilize context data, Wang et al.~\cite{DeepSD} used a large number of multiple sources as inputs of their network to predict the gap between the supply and demand of taxi in different sub-areas. Besides, some other methods~\cite{ExtremeConditionTrafficForecasting,Destination-sub-trajectory} proposed to use the recurrent neural network, like LSTM and BiLSM, to encode temporal information. With the popularity of a convolutional neural network (CNN), Yao et al.~\cite{AAAI2018} jointly modeled spatial-temporal information in a single network, and generated graph embedding additionally to extract the constant feature for each region. Though they achieved a great success in some traffic prediction fields, they neglect the discriminative temporal information hidden in time intervals and encoded sequential data without special consideration, which will both be tackled in our proposed method.

Though deep learning methods have been successful in many areas, most of them require a large amount of annotated data to be optimized. Meta Learning methods~\cite{GradientDescentbyGradientDescent,learnet,MAML,TAFENet}, however, exist to help relieve such a strict requirement by proposing more general training models that can be adjusted well to new tasks with a few new samples. Especially, Bertinetto et al. proposed a siamese-like network to receive image pairs and enforce one sub-network to generate learning weights directly for another one~\cite{learnet}. Similarly, the TAFE-Net proposed by Wang et al.~\cite{TAFENet} successfully generates weights for both convolutional and fully connected layers to another network. Inspired by such a weight generating strategy, in our work, we also explore the possibility to apply it to the demand prediction tasks, hoping to adjust our model with more adapted parameters captured by external knowledge hidden in our specific sequential data.

\section{Preliminaries}
In this section, we first introduce some basic conceptions in BSS and then formulate our demand prediction problem mathematically.

we process stations as matrices according to the geographical regions because the distribution of them is initially matrix-like

Following the definition of\cite{AAAI2018} and\cite{deepstn+}, we denote $S=\{s_1,s_2,…,s_N\}$ as the set of all stations in which the number of orders needs to be predicted, where $N$ is the total number of stations used in our dataset. These stations are further converted into a matrix $M\in \{s_n\}^{i\times j}$ where $N=i\times j$, according to the geographical distribution of these stations. For temporal information, suppose each day can be segmented into $H$ time intervals and there are $D$ days in a dataset, we define $T=\{t_{0,0},t_{1,0},…,t_{H-1,D-1}\}$ as the set of whole time intervals. Given the above definitions, we further formulate the following conceptions.

\textbf{Rental order:} A rental order $A$ can be defined as $\langle A.s,A.t\rangle$ that contains the station where people rent their bikes and the corresponding start time interval. We represent it as $\langle A.s,A.t\rangle$ with a \textit{tuple} structure where $s$ is the station and $t$ denotes the interval.

\textbf{Return order:} Similarly, a return order $R$ can be also defined as $\langle R.s,R.t \rangle$ in which $s$ and $t$ correspond to the same meaning in $A$. 

\textbf{Rental/Return demand:} The rental and return demand in one station $n$ and time interval $t_{h,d}$ are both defined as the total number of rental/return orders during that time and location, which can be denoted as $m_A/m_R$. Therefore, when dealing with all BSS stations, we set $M_A^t/M_R^t\in \mathbb{N}^{i\times j}$ as matrices with each element representing the demand of each station. Furtherly, demand matrices for all time intervals can be defined as $M_A$ and $M_R$ respectively.

\textbf{Demand:} With all definitions above, we finally concatenate two demand matrices, $M_A^t$ and $M_R^t$, together as joints input $M^t\in \mathbb{N}^{2\times i\times j}$ for our proposed network in time interval $t$. As shown in Figure \ref{fig:demand_matrix}, our demand matrix has two channels representing rental and return demands respectively. Each grid is one station and the corresponding color describes the number of orders. 

\textbf{Demand Prediction:} Given the sequential data from the beginning time to the current, demand prediction aims to predict the data in the future one time step or several steps. Especially, for the BSS demand prediction, we denote it as
\begin{equation}
    M^t= \mathcal{F}(\left\{M^{t-L},…,M^{t-2},M^{t-1}\right\}\mid \mathbb{P})
\end{equation}
where $L$ is the length of the input sequence, $\mathbb{P}$ represents some additional information that can help for prediction tasks as prior knowledge, like the spatial connection among stations~\cite{AAAI2018} and different time intervals in a day in our method.

\begin{figure*}[t]
    \centering
    \includegraphics[width=0.78\textwidth]{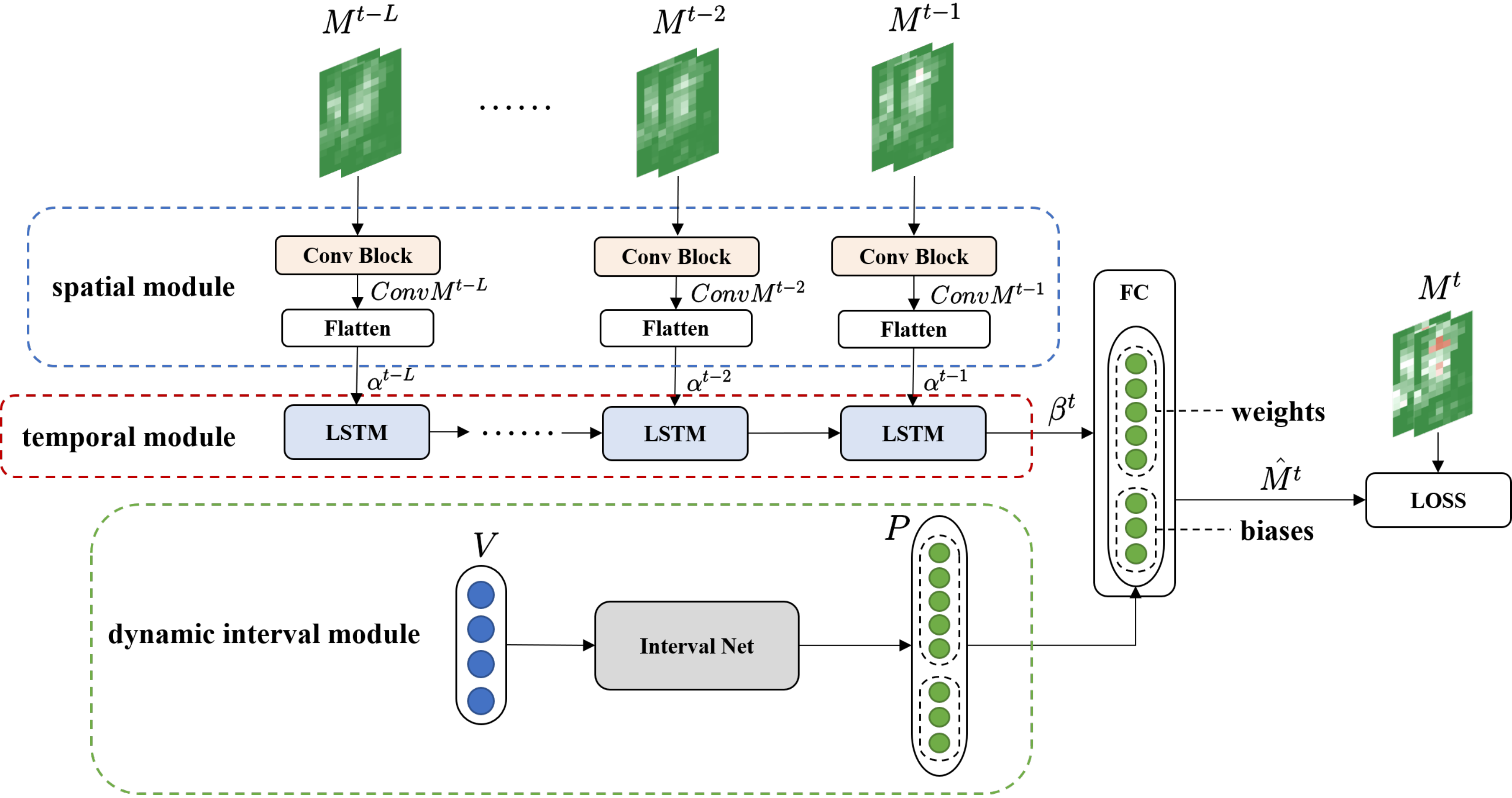}
    \caption{The Architecture of STDI-Net. The spatial module uses Conv Blocks to capture the spatial feature among stations. The Conv Block consisted of a convolutional layer and residual units. Flatten layers are used to transform the output of Conv Blocks to vectors. The temporal module uses an LSTM model to extract temporal information. The dynamic interval module takes different time intervals as inputs to generate the learnable parameters (weights and biases) for the fully connected layer.}
    \label{fig:model}
\end{figure*}

\section{Proposed Spatial-Temporal Dynamic Interval Network}
In this section, we provide the details of our proposed Spatial-Temporal Dynamic Interval Network (STDI-Net) for the demand prediction task of BSS. We first talk about our spatial-temporal module separately and then introduce the dynamic interval module which generates different parameters for the network based on time intervals in a day. Figure~\ref{fig:model} shows the overview architecture of our model.

\subsection{Spatial Module}
\begin{figure}[t]
    \centering
    \includegraphics[width=0.25\textwidth]{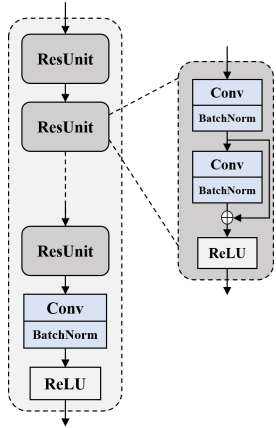}
    \caption{Internal structure of Conv Block}
    \label{fig:conv}
\end{figure}

The spatial module of the network aims to extract the joint features of all stations in each demand matrix. For each data node in one sequential input, we apply a residual convolutional block to operate on it. Inspired by~\cite{ResidualNet} that proposed the residual link to solve problems brought by very deep networks, like the vanishing gradient problem, we utilize a similar idea in our spatial module. With a concatenation between different levels of layers, the block can not only extract more abstracted representations of the demand matrix in a deep layer but also consider context information connected through different layers from the sparse input as the number of orders to the compact spatial relationships among different stations. More details are shown in Figure~\ref{fig:conv} and the process $\mathcal{F}_{s}$ can be denoted as
\begin{equation}
    \begin{split}
        &X_1 = X_0 * W_1 + b_1\\
        &X_2 = X_1 * W_2 + b_2\\
        &X_3 = f(X_1 + X_2)
    \end{split}
\end{equation}
where $X_0\in \mathbb{R}^{c_0\times i\times j}$ denotes the input of a ResUnit. $X_1\in \mathbb{R}^{c_1\times i\times j}$ and $X_2\in \mathbb{R}^{c_2\times i\times j}$ are the outputs of the first and second convolutional layers in the ResUnit respectively. $X_3\in \mathbb{R}^{c_3\times i\times j}$ represents the output of the ResUnit. The $f(\cdot)$ denotes the non-linear activation function like $ReLU$. $W_1$, $b_1$, $W_2$, and $b_2$ represent the weights and biases of the first and second convolutional layers in the ResUnit separately.

To further consider that matrices in each sequential data serves different roles based on their indexes, we create multiple independent Conv Blocks with the same structure and each of the block is responsible for one corresponding demand matrix. We denote the process as
\begin{equation}
    ConvM^l = \mathcal{F}_{s}^{l}(M^l), \, l\in{t-L,...,t-2,t-1}
\end{equation}
where $M^l\in \mathbb{R}^{2\times i\times j}$ is the two-channel demand matrix as original input on time interval $l$ and $ConvM^l \in \mathbb{R}^{c\times i \times j}$ is the output from $M^l$ operated by the Conv Block $\mathcal{F}_{s}^{l}$. $l$ represents the index of both sequential inputs and Conv Blocks, and $L$ denotes the length of the input sequence. Therefore, the number of different convolutional blocks is equal to the number of intervals in a sequential input. Each block captures the discriminative information hidden in the indexes of the data.

After the convolutional operation, we apply flatten layers to transform $ConvM^l$ that outputs from Conv Block $\mathcal{F}_{s}^{l}$ to a feature vector $\alpha_l \in \mathbb{R}^{cij}$, where $c$ is the number of channels of the output matrix. The whole output $S_t\in \mathbb{R}^{l\times cij}$ represents all features extracted from temporal demand matrices separately, which can be denoted as:
\begin{equation}
    S_t = [\alpha_l|l=\text{t-L,…,t-2,t-1}]
\end{equation}

\subsection{Temporal Module}
Since the transportation data is a type of time series, we apply the temporal module to capture the temporal dependence of the sequential demand matrices. In the task of sequence learning, Recurrent Neural Networks (RNN) have achieved good results~\cite{seq2seq}. The incorporation of Long Short-Term Memory (LSTM) overcomes the shortage of traditional recurrent networks that learning long-term dependencies is difficult~\cite{GradientFlowinRecurrentNets}. Some previous works~\cite{Origin-Destination-Demand,AAAI2018} have proved the great performance of LSTM in processing traffic sequential data. To follow them, we apply the LSTM network for the BSS sequential data in our temporal module.

Briefly speaking, LSTM maintains a memory cell $c_t$ to accumulate the previous sequence information. Specifically, at time $t$, given an input $x_t$, the LSTM uses an input gate $i_t$ and a forget gate $f_t$ to update its memory cell $c_t$, and uses an output gate $o_t$ to control the hidden state $h_t$. The formulation is defined as follows:
\begin{equation}
    \begin{split}
        &i_t=\sigma(W_{ii}x_t + b_{ii} + W_{hi}h_{t-1} + b_{hi})\\
        &f_t=\sigma(W_{if}x_t + b_{if} + W_{hf}h_{t-1} + b_{hf})\\
        &g_t=\tanh(W_{ig}x_t + b_{ig} + W_{hg}h_{t-1} + b_{hg})\\
        &o_t=\sigma(W_{io}x_t + b_{io} + W_{ho}h_{t-1} + b_{ho})\\
        &c_t=f_t\circ c_{t-1} + i_t\circ g_t\\
        &h_t=o_t\circ \tanh(c_t)
    \end{split}
\end{equation}
where $\circ$ denotes the Hadamard product, and $\sigma$ represents the sigmoid function. $W_{\alpha\beta},b_{\alpha\beta}(\alpha\in(i,h),\beta\in(i,f,g,o))$ are the learnable parameters of the LSTM while $c_{t-1}$ and $h_{t-1}$ are the memory cell state and the hidden state at time $t-1$. Please refer to~\cite{LSTM,GradientFlowinRecurrentNets} for more details.

In our model, the LSTM net takes $S_t$ as input, which is the output of the spatial module. We use $\beta^t\in \mathbb{R}^d$ to represent the output of the LSTM net in our temporal module.

\subsection{Dynamic Interval Module}
Though the sequential demand data of BSS holds a kind of trend during the day, their changes will vary according to different time intervals. Therefore, we propose a dynamic interval module that extracts temporal information from each hour and then apply them to influence the learning strategies of the main spatial-temporal network directly.

To encourage such a learning mode, some meta-learning methods~\cite{learnet,TAFENet} have been proposed to create a siamese-like network in which one network is responsible for generating learning weights for another. Inspired by these advanced works, we also apply a similar network structure to map (time) to be directly the learning weights of the top fully connected layer in the main network. 

In our module, for the input number of hours ranging from $0$ to $23$, we first use $GloVe$~\cite{GloVe} to embed the numbers into feature vectors $V_t\in \mathbb{R}^h$. After that, our Interval Net in the module transforms embedding vectors to features whose dimension is the same as the learnable parameters in the fully connected layer of the main network, including weights and biases. The generated vectors are then directly assigned to be the values in the fully connected layer, and the Dynamic Interval Module participates in the back-propagation process in an end-to-end manner.

\begin{figure}[t]
    \centering
    \includegraphics[width=0.38\textwidth]{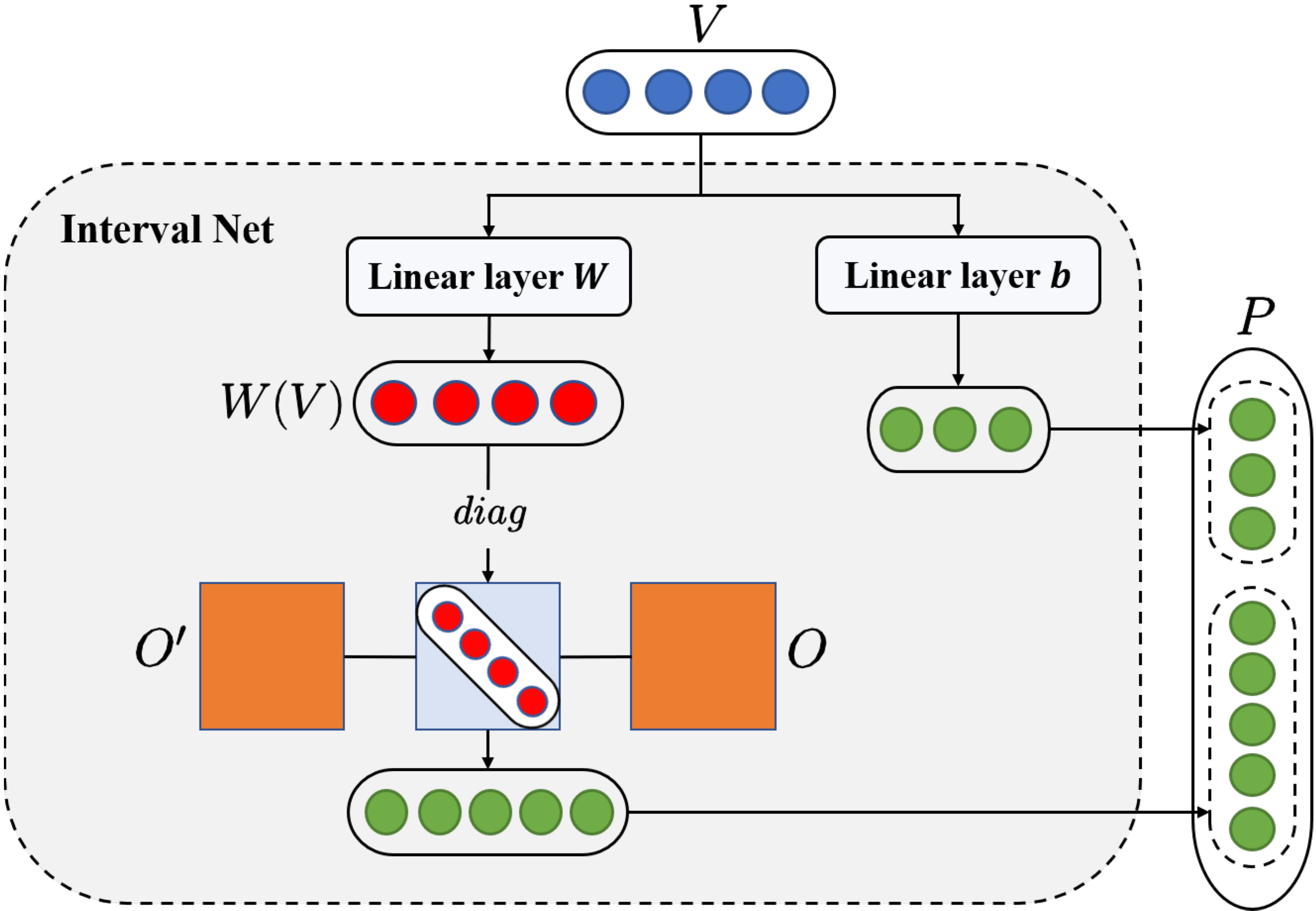}
    \caption{Internal structure of Interval Net}
    \label{fig:interval}
\end{figure}

However, it is too difficult and too large for parameters in Interval Net to learn, since the parameters space of the Interval Net grows quadratically with the number of the output units. Following~\cite{learnet}, we construct a factorized representation of the output weights that is decomposed of 2 operating matrices and a diagonal matrix as Figure~\ref{fig:interval} shows, which is analogous to the Singular Value Decomposition. By this way, the parameters in the Interval Net needed to be learned only grow linearly with the number of output units. The whole process can be formulated as

\begin{equation}
    W_{FC} = O'\,diag(W(V))\,O
\end{equation}
where $W_{FC}\in \mathbb{R}^{k\times d}$ is the generated weights for the fully connected layer. $W(V)\in \mathbb{R}^a$ represents the output vector of the Linear layer $W$ in Interval Net while $diag(\cdot)$ is the diagonal operating to transform the vector $W(V)$ to a diagonal matrix. As a consequence, the net only needs to generate low-dimensional parameters for each time interval. In addition, two matrices $O\in \mathbb{R}^{a \times d}$ and $O'\in \mathbb{R}^{k \times a}$, where $k=2\times i\times j$, project $diag(W(V))$ again to keep the same dimension with the fully connected layer.

Similarly, biases of the fully connected layer are also generated as following:
\begin{equation}
    b_{FC} = b(V)
\end{equation}
where $b_{FC}$ represents the generated biases for the fully connected layer. $b(V)\in \mathbb{R}^k$ denotes the output vector of the linear layer $b$ in Interval Net. After the above operation, we obtain $\langle W_{FC},b_{FC}\rangle$ as the parameters $P$ in Figure~\ref{fig:model} of the fully connected layer (FC).

To get the final results, the fully connected layer takes the output of temporal module $\beta^t\in \mathbb{R}^d$ as input for the time interval $t$. As we mentioned, $P$ consists of the weights $W_{FC}\in \mathbb{R}^{k\times d}$ and biases $b_{FC} \in \mathbb{R}^k$ where $k=2\times i\times j$. Therefore, the formulation of the layer can be expressed as follows:
\begin{equation}
    \hat{M_t}=f(W_{FC} \beta^t + b_{FC})
\end{equation}
where the $f(\cdot)$ denotes the non-linear activation function of prediction layer. $\hat{M_t} \in \mathbb{R}^k$ represents the predicted demand matrix of the ground truth $M_t$.

\subsection{Implementation Details}
In the experiments, we set the length of the input sequence $L$ to 3. In the spatial module, each Conv Block has 2 ResUnits with the same structure. That is, it contains 2 convolutional layers with each layer followed by a batch normalization (BN)~\cite{BN} and a residual link. All the convolutional layers in the Conv Block have 32 filters. The size of each filter is set to $3\times3$ with $stride=1$. In the temporal module, the LSTM net has 1 hidden layer with 1024 neurons. The activation functions used in the fully connected layer and Conv Blocks are $ReLU$ while $LeakyReLU$ is used as the activation function at the linear layers in the dynamic interval module. We optimize our model via Adam~\cite{Adam} optimization by minimizing the Mean Squared Error (MSE) loss between the predicted result and the ground truth. The learning rate and the weight decay are set to $10^{-3}$ and 5e-5 respectively. For the training data, 90\% of it is for training and the remaining 10\% is chosen as a validation set for early-stop. We implement our network with Pytorch~\cite{pytorch} and train it for 200 epochs on 2 NVIDIA 1080Ti GPUs.

\section{Experiment}
\subsection{Dataset}
In the paper, we use the NYC Bike dataset in 2014, from Apr. 1st to Sept. 30th. We treat the data for the last 10 days as the testing data and others as training data. We set one hour as the length of a time interval. The total number of orders and time intervals in the dataset are 5,359,944 and 4,392 respectively. And the number of stations used in the dataset is 128. The dataset can be collected from the website of Citi-Bike system\footnote{https://www.citibikenyc.com/system-data}. 


\subsection{Evaluation Metric}
We use Rooted Mean Square Error (RMSE) and Mean Absolute Error (MAE) as the metrics to evaluate the performance of our model and the baselines, which are defined as:

\begin{equation}
    RMSE=\sqrt{\frac{1}{z}\sum_{i}(y_i-\hat{y_i})^2}
\end{equation}

\begin{equation}
    MAE=\frac{1}{z}\sum_{i=1}^{z}|y_i-\hat{y_i}|
\end{equation}

where $\hat{y_i}$ and ${y_i}$ denote the predicted value and ground truth respectively, and ${z}$ is the number of all predicted values. 

\subsection{Baselines}
We compare our STDI-Net with the following seven baselines:
\begin{itemize}
    \item \textbf{Historical average (HA)}: Historical Average (HA) predicts the future demand by averaging the historical demands.
	\item \textbf{Auto-regressive integrated moving average (ARIMA)}: Auto-Regression Integrated Moving Average (ARIMA) is a well-known model used for time series prediction.
	\item \textbf{Lasso regression (Lasso)}: Lasso regression is a linear regression method with $L_1$ regularization.
    \item \textbf{Ridge regression (Ridge)}: Ridge regression is a linear regression method with $L_2$ regularization.
	\item \textbf{Multiple layer perception (MLP)}: MLP is a neural network with four hidden layers. The number of hidden units are 256, 256, 128, 128 respectively.
	The MLP predicts the demand matrix $M^t$ by taking a sequence of the previous $l$ demand matrix $[M^{t-l},…,M^{t-2},M^{t-1}]$ as input.
	\item \textbf{ST-ResNet}~\cite{AAAI2017}: ST-ResNet is a CNN-based model with residual blocks for traffic prediction, which used multiple CNN components to extract features from the historical data sequence.
	\item \textbf{DMVST-Net}~\cite{AAAI2018}: DMVST-Net is a deep learning model which based on CNN and LSTM for taxi demand prediction. It also contains graph embedding to capture similar demand patterns among regions.
	\item \textbf{DeepSTN+}~\cite{deepstn+}: DeepSTN+ is a deep learning-based convolutional model for crowd flow prediction, which contains long range spatial dependence modeling, POI-based spatial information capturing, and a fusion mechanism for features extracted from different aspects.
\end{itemize}

\begin{table}

\caption{Comparison with baselines.}
\centering
\begin{tabular}{l<{\centering}|p{40pt}<{\centering}|p{40pt}<{\centering}}
\toprule
Method & RMSE & MAE\\
\hline
\hline
Historical average & 10.7308 & 5.8374\\
ARIMA & 10.4773 & 4.7005\\
Lasso regression & 8.4947 & 3.6799\\
Ridge regression & 8.4699 & 3.6984\\
Multiple layer perception & 7.1888 & 3.3388\\
\hline
ST-ResNet & 5.1249 & 2.7206\\
DMVST-Net & 5.0595 & 2.3423\\
DeepSTN+ & 4.9060 & 2.4269\\
\hline
STDI-Net & \textbf{4.6339} & \textbf{2.1946}\\
\bottomrule
\end{tabular}
\label{tab:baselines}
\end{table}

\subsection{Comparison with Baselines}
Table~\ref{tab:baselines} shows the testing results of our proposed model and baselines on the dataset. We can see that our STDI-Net achieves the lowest RMSE and MAE(4.6339 and 2.1946) among all the competing methods. The HA and ARIMA perform poorly, as they only consider the historical demand values for prediction. Because of the consideration of more context relationships among sequence, the linear regression methods (Lasso and Ridge) perform better than the above two methods. However, they do not extract more spatial-temporal information for prediction. The MLP further extracts features from the sequence and performs better than the above methods. However, the MLP does not model spatial or temporal dependency. The ST-ResNet achieves 5.1249 and 2.7206 for RMSE and MAE which is better than MLP due to the extracting of spatial features. Compared with ST-ResNet, DMVST-Net extracts joint spatial-temporal feature and similar demand patterns among regions, which further improve its performance for prediction. Compared with previous methods, DeepSTN+ explores spatial correlations from different aspects to reduce the prediction error. However, it doesn't consider about the influence of different time intervals. Our model further contributes a dynamic interval module which further improves the performance.

\subsection{Comparison with Modules Combinations}
Our full model consists of three modules for three types of information modeling. To explore the influence of different modules combinations on the task, we combine them and implement the following networks:
\begin{itemize}
    \item \textbf{Spatial module + FC}: This network contains the spatial module of our proposed model and a fully connected layer. This network only extracts spatial features for prediction.
    \item \textbf{Temporal module + FC}: This network only uses the temporal module of our proposed model to capture the temporal information, and a fully connected layer is used to output the predicted results.
    \item \textbf{Spatial module + Temporal module + FC}: This method is the combination of the spatial module, temporal module, and a fully connected layer. In this method, we model joint spatial-temporal information without considering the influence of different time intervals.
    \item \textbf{Spatial module + Dynamic Interval module}: In this network, we combine the spatial module and the dynamic interval module of our proposed model, to capture spatial information, and the dynamic mappings for different time intervals.
    \item \textbf{Temporal module + Dynamic Interval module}: For this network, we use the temporal module and the dynamic interval module of our proposed model. This network models the temporal information, and generates the dynamic mappings for different time intervals.
    \item \textbf{STDI-Net}: Our proposed model, which models joint spatial-temporal information, and generates dynamic mappings for different time intervals.
\end{itemize}

Table~\ref{tab:combination} shows the results of the test. The RMSE and MAE of the spatial module + FC are 5.6558 and 2.6218 respectively, while that of the spatial module + dynamic interval module are 4.9077 and 2.3457. The results of the temporal module + FC achieve 5.2614 and 2.3914 while the RMSE and MAE of the temporal module + dynamic interval module are 4.7788 and 2.2582 respectively. We can see that compared with separate spatial or temporal module + fully connected layer, the performance of the combination with the dynamic interval module improves significantly. Furthermore, the spatial module + temporal module + FC achieves the results of 5.0832 and 2.3476, which are worse than that of our complete model. The results show that our dynamic interval module improves the performance significantly.

\begin{table}
\centering
\caption{Comparison with Different Modules Combinations}
\begin{tabular}{l|p{35pt}<{\centering}|p{35pt}<{\centering}}
\toprule
Method & RMSE & MAE\\
\hline
\hline
Spatial + FC & 5.6558 & 2.6218\\
Temporal + FC & 5.2614 & 2.3914\\
Spatial + Temporal + FC &  5.0832 & 2.3476\\
Spatial + Dynamic Interval & 4.9077 & 2.3457\\
Temporal + Dynamic Interval & 4.7788 & 2.2582\\
\hline
STDI-Net & \textbf{4.6339} & \textbf{2.1946}\\
\bottomrule
\end{tabular}
\label{tab:combination}
\end{table}

\subsection{Comparison with Variants of Our Model}

The above experiments show that our proposed dynamic interval module achieves a good result in the demand prediction of BSS. However, we have not proved the rationality of the parameters-generated mode in the dynamic interval module. Besides, we also need to evaluate the effectiveness of the time-specific convolutional layers in our spatial module. In addition, the advantage of using $GloVe$ need to be proved by comparing with the model that embed time intervals into vectors without the pre-trained $GloVe$. To address these two questions, we construct the following three variants of our proposed model:
\begin{itemize}
    \item \textbf{STDI-Net-fusion}: In this network, we apply a Linear layer in the Interval Net to transform the interval embedding vector to a feature, and then we concatenate it with the output of the temporal module. After that, a fully connected layer is used to output the predicted results.
    \item \textbf{Unified-Spatial Net}: This network is the variant of our proposed spatial module, which is used to evaluate the performance of applying the same filters in different temporal indexes. This model Net applies unified filters for each index of the sequence in all convolutional layers, and a fully connected layer is used after convolutional layers. Note that, in the Unified-spatial Net, we use the same Conv Blocks structure as our proposed STDI-Net.
    \item \textbf{STDI-Net-embedding}: In this model, we apply a learnable embedding layer to embed the hours' number instead of using the pre-trained $GloVe$ to embed them.
\end{itemize}

\begin{table}
\centering
\caption{Comparison with Variants of Our Model}
\begin{tabular}{l|p{35pt}<{\centering}|p{35pt}<{\centering}}
\toprule
Method & RMSE & MAE\\
\hline
\hline
Unified-Spatial Net & 6.1493 & 2.9533\\
Spatial module + FC & \textbf{5.6558} & \textbf{2.6218}\\
STDI-Net-fusion & 4.8149 & 2.2995\\
STDI-Net-embedding & 4.6154 & 2.1783\\
STDI-Net & \textbf{4.6339} & \textbf{2.1946}\\
\bottomrule
\end{tabular}
\label{tab:Variants}
\end{table}

Table~\ref{tab:Variants} shows the results of the above three variants of our model. We can see that our spatial module + FC (5.6558 and 2.6218 for RMSE and MAE) outperforms Unified-Spatial Net (6.1493 and 2.9533 for RMSE and MAE), that means, our proposed time-specific convolution layers perform better than applying same convolutional filters in different temporal indexes. Otherwise, STDI-Net-fusion achieves 4.8149 for RMSE and 2.2995 for MAE, which are worse than our STDI-Net (4.6339 and 2.1946 respectively). Therefore, our parameters-generated mode is better than the fusion way.

Due to applying a trainable embedding layer instead of using a pre-trained model ($GloVe$), the STDI-Net-embedding (4.6154 and 2.1783) has more learnable parameters than STDI-Net (4.6339 and 2.1946). Therefore it can perform better than our STDI-Net. However, its performance has not improved significantly (0.4$\%$ and 0.7$\%$ for RMSE and MAE respectively) with additional parameters. That means, our STDI-Net can perform almost as well as STDI-Net-embedding with less parameters than it. To reduce the number of learnable weights, we apply $GloVe$ to embed hours instead of using an additional embedding layer to embed them.

\section{Conclusion and Discussion}
In this paper, we propose a novel deep learning-based method for demand prediction of Bike Sharing System (BSS). Our model considers the extraction of joint spatial-temporal feature and time-specific convolutional layers with residual links. Furthermore, we contribute a dynamic interval module to include the factor that different time intervals have a strong influence on demand prediction in BSS by generating different feature mappings for different time intervals. We evaluate our model on the NYC Bike dataset, and the results show that our model significantly outperforms the competing baselines. In the future, we will consider some other features to further improve the performance of our model, such as meteorology data, holiday data. And we will consider the more dependent relationship of stations, such as use Graph Convolutional Network (GCN) to extract the spatial feature among stations.

\bibliography{paper}

\begin{thebibliography}{}

\bibitem[\protect\citeauthoryear{Andrychowicz \bgroup et al\mbox.\egroup
  }{2016}]{GradientDescentbyGradientDescent}
Andrychowicz, M.; Denil, M.; Gomez, S.; Hoffman, M.~W.; Pfau, D.; Schaul, T.;
  Shillingford, B.; and De~Freitas, N.
\newblock 2016.
\newblock Learning to learn by gradient descent by gradient descent.
\newblock In {\em NeurIPS}.

\bibitem[\protect\citeauthoryear{Bertinetto \bgroup et al\mbox.\egroup
  }{2016}]{learnet}
Bertinetto, L.; Henriques, J.~a.~F.; Valmadre, J.; Torr, P.; and Vedaldi, A.
\newblock 2016.
\newblock Learning feed-forward one-shot learners.
\newblock In {\em NeurIPS},  523--531.

\bibitem[\protect\citeauthoryear{Chiang, Hoang, and
  Lim}{2015}]{GaussianMixtureModel}
Chiang, M.-F.; Hoang, T.-A.; and Lim, E.-P.
\newblock 2015.
\newblock Where are the passengers?: A grid-based gaussian mixture model for
  taxi bookings.
\newblock In {\em SIGSPATIAL}.

\bibitem[\protect\citeauthoryear{DeMaio}{2009}]{bikesharing1}
DeMaio, P.
\newblock 2009.
\newblock Bike-sharing: History, impacts, models of provision, and future.
\newblock {\em Journal of Public Transportation} 12(4):41--56.

\bibitem[\protect\citeauthoryear{Finn, Abbeel, and Levine}{2017}]{MAML}
Finn, C.; Abbeel, P.; and Levine, S.
\newblock 2017.
\newblock Model-agnostic meta-learning for fast adaptation of deep networks.
\newblock In {\em ICML}.

\bibitem[\protect\citeauthoryear{He \bgroup et al\mbox.\egroup
  }{2016}]{ResidualNet}
He, K.; Zhang, X.; Ren, S.; and Sun, J.
\newblock 2016.
\newblock Deep residual learning for image recognition.
\newblock In {\em CVPR}.

\bibitem[\protect\citeauthoryear{Hochreiter and Schmidhuber}{1997}]{LSTM}
Hochreiter, S., and Schmidhuber, J.
\newblock 1997.
\newblock Long short-term memory.
\newblock {\em Neural computation} 9(8):1735--1780.

\bibitem[\protect\citeauthoryear{Informatik \bgroup et al\mbox.\egroup
  }{2003}]{GradientFlowinRecurrentNets}
Informatik, F.; Bengio, Y.; Frasconi, P.; and Schmidhuber, J.
\newblock 2003.
\newblock Gradient flow in recurrent nets: the difficulty of learning long-term
  dependencies.
\newblock {\em A Field Guide to Dynamical Recurrent Neural Networks}.

\bibitem[\protect\citeauthoryear{Ioffe and Szegedy}{2015}]{BN}
Ioffe, S., and Szegedy, C.
\newblock 2015.
\newblock Batch normalization: Accelerating deep network training by reducing
  internal covariate shift.
\newblock In {\em ICML}.

\bibitem[\protect\citeauthoryear{Kingma and Ba}{2014}]{Adam}
Kingma, D.~P., and Ba, J.
\newblock 2014.
\newblock Adam: A method for stochastic optimization.
\newblock {\em arXiv preprint arXiv:1412.6980}.

\bibitem[\protect\citeauthoryear{Krizhevsky, Sutskever, and
  Hinton}{2012}]{ImageNet}
Krizhevsky, A.; Sutskever, I.; and Hinton, G.~E.
\newblock 2012.
\newblock Imagenet classification with deep convolutional neural networks.
\newblock In {\em NeurIPS},  1097--1105.

\bibitem[\protect\citeauthoryear{{Moreira-Matias} \bgroup et al\mbox.\egroup
  }{2013}]{DemandStreamingData}
{Moreira-Matias}, L.; {Gama}, J.; {Ferreira}, M.; {Mendes-Moreira}, J.; and
  {Damas}, L.
\newblock 2013.
\newblock Predicting taxi–passenger demand using streaming data.
\newblock {\em IEEE Transactions on Intelligent Transportation Systems}
  14(3):1393--1402.

\bibitem[\protect\citeauthoryear{Pan, Demiryurek, and
  Shahabi}{2012}]{realworld}
Pan, B.; Demiryurek, U.; and Shahabi, C.
\newblock 2012.
\newblock Utilizing real-world transportation data for accurate traffic
  prediction.
\newblock In {\em ICDM}.

\bibitem[\protect\citeauthoryear{Paszke \bgroup et al\mbox.\egroup
  }{2019}]{pytorch}
Paszke, A.; Gross, S.; Massa, F.; Lerer, A.; Bradbury, J.; Chanan, G.; Killeen,
  T.; Lin, Z.; Gimelshein, N.; Antiga, L.; et~al.
\newblock 2019.
\newblock Pytorch: An imperative style, high-performance deep learning library.
\newblock In {\em NeurIPS}.

\bibitem[\protect\citeauthoryear{Pennington, Socher, and Manning}{2014}]{GloVe}
Pennington, J.; Socher, R.; and Manning, C.
\newblock 2014.
\newblock {G}love: Global vectors for word representation.
\newblock In {\em EMNLP}.

\bibitem[\protect\citeauthoryear{{Qiu} \bgroup et al\mbox.\egroup
  }{2019}]{Origin-Destination-Demand}
{Qiu}, Z.; {Liu}, L.; {Li}, G.; {Wang}, Q.; {Xiao}, N.; and {Lin}, L.
\newblock 2019.
\newblock Taxi origin-destination demand prediction with contextualized
  spatial-temporal network.
\newblock In {\em ICME},  760--765.

\bibitem[\protect\citeauthoryear{Shaheen, Guzman, and
  Zhang}{2010}]{bikesharing2}
Shaheen, S.~A.; Guzman, S.; and Zhang, H.
\newblock 2010.
\newblock Bikesharing in europe, the americas, and asia: Past, present, and
  future.
\newblock {\em Transportation Research Record} 2143(1):159--167.

\bibitem[\protect\citeauthoryear{Shekhar and Williams}{2007}]{AdaptiveSeasonal}
Shekhar, S., and Williams, B.~M.
\newblock 2007.
\newblock Adaptive seasonal time series models for forecasting short-term
  traffic flow.
\newblock {\em Transportation Research Record} 2024(1):116--125.

\bibitem[\protect\citeauthoryear{Sutskever, Vinyals, and Le}{2014}]{seq2seq}
Sutskever, I.; Vinyals, O.; and Le, Q.~V.
\newblock 2014.
\newblock Sequence to sequence learning with neural networks.
\newblock In {\em NeurIPS},  3104--3112.

\bibitem[\protect\citeauthoryear{{Wang} \bgroup et al\mbox.\egroup
  }{2017}]{DeepSD}
{Wang}, D.; {Cao}, W.; {Li}, J.; and {Ye}, J.
\newblock 2017.
\newblock Deepsd: Supply-demand prediction for online car-hailing services
  using deep neural networks.
\newblock In {\em ICDE},  243--254.

\bibitem[\protect\citeauthoryear{Wang \bgroup et al\mbox.\egroup
  }{2019}]{TAFENet}
Wang, X.; Yu, F.; Wang, R.; Darrell, T.; and Gonzalez, J.~E.
\newblock 2019.
\newblock Tafe-net: Task-aware feature embeddings for low shot learning.
\newblock In {\em CVPR}.

\bibitem[\protect\citeauthoryear{Wu, Wang, and Li}{2016}]{UbiquitousUrbanData}
Wu, F.; Wang, H.; and Li, Z.
\newblock 2016.
\newblock Interpreting traffic dynamics using ubiquitous urban data.
\newblock In {\em SIGSPACIAL}.

\bibitem[\protect\citeauthoryear{Yao \bgroup et al\mbox.\egroup
  }{2018}]{AAAI2018}
Yao, H.; Wu, F.; Ke, J.; Tang, X.; Jia, Y.; Lu, S.; Gong, P.; Ye, J.; and
  Zhenhui, L.
\newblock 2018.
\newblock Deep multi-view spatial-temporal network for taxi demand prediction.
\newblock In {\em AAAI}.

\bibitem[\protect\citeauthoryear{Yu \bgroup et al\mbox.\egroup
  }{2017}]{ExtremeConditionTrafficForecasting}
Yu, R.; Li, Y.; Shahabi, C.; Demiryurek, U.; and Liu, Y.
\newblock 2017.
\newblock Deep learning: A generic approach for extreme condition traffic
  forecasting.
\newblock In {\em SIAM}.

\bibitem[\protect\citeauthoryear{Zhang \bgroup et al\mbox.\egroup
  }{2016}]{DeepST}
Zhang, J.; Zheng, Y.; Qi, D.; Li, R.; and Yi, X.
\newblock 2016.
\newblock Dnn-based prediction model for spatio-temporal data.
\newblock In {\em SIGSPATIAL}.

\bibitem[\protect\citeauthoryear{Zhang, Zheng, and Qi}{2017}]{AAAI2017}
Zhang, J.; Zheng, Y.; and Qi, D.
\newblock 2017.
\newblock Deep spatio-temporal residual networks for citywide crowd flows
  prediction.
\newblock In {\em AAAI}.

\bibitem[\protect\citeauthoryear{Zhao \bgroup et al\mbox.\egroup
  }{2018}]{Destination-sub-trajectory}
Zhao, J.; Xu, J.; Zhou, R.; Zhao, P.; Liu, C.; and Zhu, F.
\newblock 2018.
\newblock On prediction of user destination by sub-trajectory understanding: A
  deep learning based approach.
\newblock In {\em CIKM}, CIKM '18,  1413--1422.

\end{thebibliography}

\end{document}